%% file: conference_101719.tex
\documentclass[conference]{IEEEtran}
\IEEEoverridecommandlockouts
\usepackage{cite}
\usepackage{amsmath,amssymb,amsfonts}
\usepackage{algorithmic}
\usepackage{graphicx}
\usepackage{textcomp}
\usepackage{xcolor}
\usepackage{multirow}
\usepackage{algorithm}
\usepackage{multirow}
\usepackage{booktabs}
\def\BibTeX{{\rm B\kern-.05em{\sc i\kern-.025em b}\kern-.08em
    T\kern-.1667em\lower.7ex\hbox{E}\kern-.125emX}}
\begin{document}

\title{Interactive State Space Model with Cross-Modal Local Scanning for Depth Super-Resolution
}


\author{
	\IEEEauthorblockN{
		Chen Wu\textsuperscript{1}, 
		Ling Wang\textsuperscript{2}, 
        Zhuoran Zheng\textsuperscript{3},
		Xiangyu Chen\textsuperscript{4}\IEEEauthorrefmark{1}, 
		Jingyuan Xia\textsuperscript{1}\thanks{\IEEEauthorrefmark{1}These authors contributed equally as co-corresponding authors.}\IEEEauthorrefmark{1}, 
        Weidong Jiang\textsuperscript{1}
		and Jiantao Zhou\textsuperscript{5}
        \thanks{This work is supported by the National Natural Science Foundation of China under Grant 62576350, 62131020 and 625B2180.}} 
        \textsuperscript{1}National University of Defense Technology 
        \textsuperscript{2}The Hong Kong University of Science and Technology (Guangzhou)\\
        \textsuperscript{3}Sun Yat-sen University
        \textsuperscript{4}Institute of Artificial Intelligence (TeleAI), China Telecom
        \textsuperscript{5}University of Macau}
\maketitle

\begin{abstract}
Guided depth super-resolution (GDSR) reconstructs HR depth maps from LR inputs with HR RGB guidance. 
Existing methods either model each modality independently or rely on computationally expensive attention mechanisms with quadratic complexity, hindering the establishment of efficient and semantically interactive joint representations.
In this paper, we observe that feature maps from different modalities exhibit semantic-level correlations during feature extraction. This motivates us to develop a more flexible approach enabling dense, semantically-aware deep interactions between modalities.
To this end, we propose a novel GDSR framework centered around the Interactive State Space Model. Specifically, we design a cross-modal local scanning mechanism that enables fine-grained semantic interactions between RGB and depth features. Leveraging the Mamba architecture, our framework achieves global modeling with linear complexity.
Furthermore, a cross-modal matching transform module is introduced to enhance interactive modeling quality by utilizing representative features from both modalities. 
Extensive experiments demonstrate competitive performance against state-of-the-art methods.
\end{abstract}


\section{Introduction}
Depth maps capture geometric distance relationships within a scene and are widely employed in various visual understanding tasks, such as autonomous driving~\cite{wang2019pseudo,yurtsever2020survey}, object recognition~\cite{chowdhary20193d}, and virtual reality~\cite{bonetti2018augmented,burdea2003virtual}. However, due to limitations in sensor technology and imaging conditions, depth images acquired by depth-sensing cameras often suffer from non-ideal and often unknown degradations~\cite{SGNet}. To mitigate hardware constraints and reduce costs, some studies have explored the use of RGB images to guide the super-resolution of depth maps~\cite{riegler2016atgv,song2016deep,voynov2019perceptual}. With promising results in low-level tasks~\cite{wu2025ultra,wang2025dap,wu2025adaptive,chen2025mixnet,RSAG,SGNet,DRMPNet}, recent deep learning research in depth image super-resolution has also become predominantly learning-based. For instance, SUFT~\cite{SUFT} introduces a symmetric uncertainty method that adaptively selects informative regions from RGB images to aid depth reconstruction while minimizing the influence of irrelevant textures. In a different vein, RSAG~\cite{RSAG} develops a recursive structure attention mechanism to progressively restore high-frequency structural details in depth maps. Furthermore, SGNet~\cite{SGNet} incorporates spectral and gradient maps as auxiliary cues to explicitly enhance edge-aware features and improve the recovery of structural boundaries. Meanwhile, DRMPNet~\cite{DRMPNet} employs a cross-attention architecture to facilitate feature alignment and fusion between depth and RGB modalities, enabling more effective cross-modal representation learning. However, these methods typically handle each modality independently and then perform joint processing through simple inter-modal feature interactions, such as element-wise addition, multiplication, or concatenation. On the other hand, some methods employ complex attention mechanisms for joint modeling, resulting in quadratic growth in computational complexity, which hinders the establishment of efficient and semantically interactive joint representations.

\begin{figure}[t!]
    \centering
    \includegraphics[width=.88\linewidth]{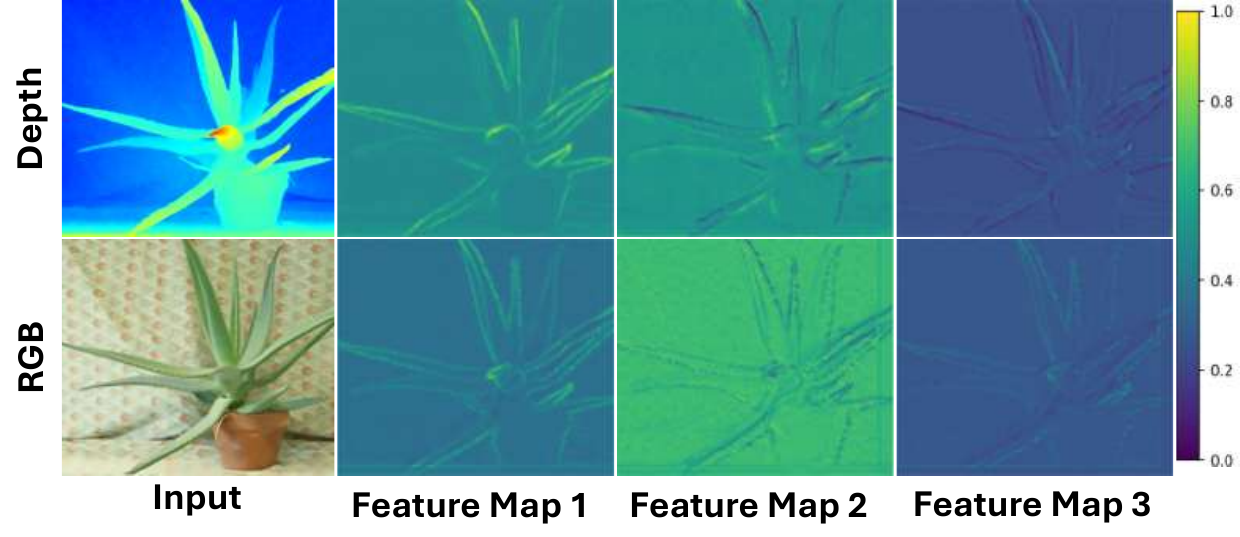}
    \caption{We observed that certain feature maps between different modalities exhibit strong semantic correlations, prompting us to explore joint interactive modeling to achieve high-quality output results.}
    \label{fig:motivation}
\end{figure}

Recently, the Mamba~\cite{Mamba} has garnered significant attention in the low-level domain, with its linear complexity providing a foundation for efficient model processing, while its core scanning mechanism offers a flexible approach to global modeling~\cite{peng2025directing,he2024multi,xia2024s3mamba,xu2026fusion}. In this paper, as shown in Fig.~\ref{fig:motivation}, we observe that different modalities exhibit strong semantic correlations in their feature maps after feature extraction. This inspires us to integrate these semantically related feature maps during global modeling to achieve deep inter-modal interactions at the semantic level, thereby reducing information decay when capturing long-range dependencies. To this end, we propose a novel GDSR framework, which consists of two uniquely designed components: the interactive state space model and the cross-modal matching transform module. The former optimizes the scanning mechanism in Mamba to form a cross-modal local scanning approach, taking into account fine-grained semantic interactions between RGB and depth features while modeling long-range dependencies. Notably, we apply a patchify operation to the feature maps, introducing a local inductive bias~\cite{huang2024localmamba}. The latter explicitly aligns and fuses similar semantic features across different modalities, thereby enhancing the representation of each modality through feature interaction and integration. Through extensive experimental comparisons with existing state-of-the-art methods on multiple benchmarks, we demonstrate that our approach achieves competitive performance.

\begin{itemize}
    \item We propose an efficient global modeling framework based on the interactive state space model, which introduces the linear complexity Mamba architecture to GDSR tasks for the first time.
    \item We design a set of collaboratively working cross-modal interaction mechanisms, including cross-modal local scanning and cross-modal matching transform module, which achieve deep and efficient modality interaction from both semantic association and feature matching perspectives.
    \item Extensive experiments demonstrate the superiority of our approach over state-of-the-art methods in terms of accuracy and generalization on three datasets.
\end{itemize}

\begin{figure*}[t!]
    \centering
    \includegraphics[width=.85\linewidth]{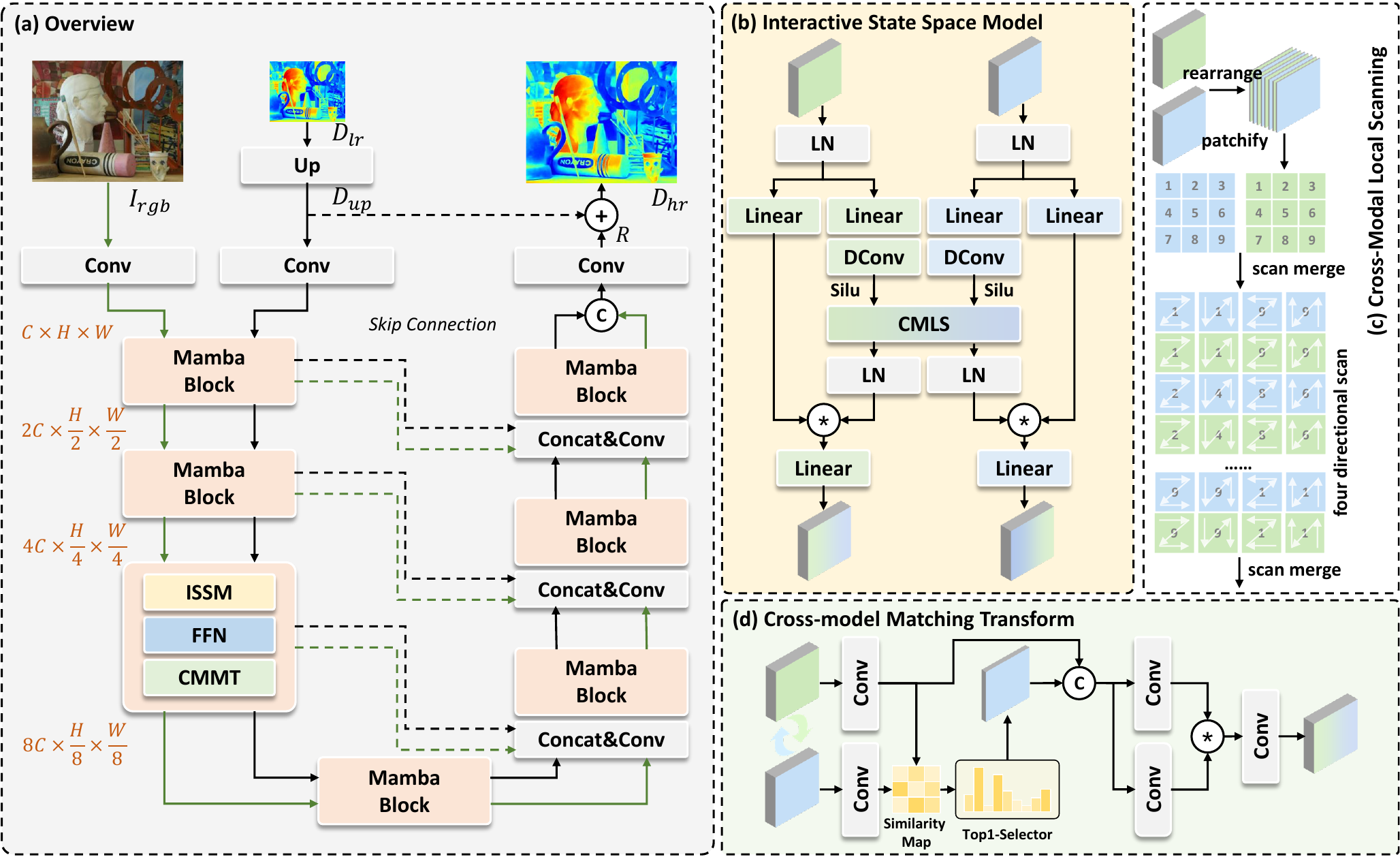}
    \caption{The overview of our proposed method highlights its core components: an ISSM with a CMLS mechanism and a CMMT module.}
    \label{fig:framework}
\end{figure*}

\section{Method}
\subsection{Overview}
As illustrated in Fig.~\ref{fig:framework} (a), given a LR depth map $\boldsymbol{D}_{lr} \in \mathbb{R}^{1 \times H \times W}$ and a HR RGB image $\boldsymbol{I}_{RGB} \in \mathbb{R}^{3 \times sH \times sW}$, where $H \times W$ denotes spatial dimensions and $s$ is the scaling factor, the LR depth map is first upsampled to match the spatial resolution of the HR RGB image, yielding an initial HR depth estimate $\boldsymbol{D}_{up} \in \mathbb{R}^{1 \times sH \times sW}$. The aligned inputs are then projected into a high-dimensional feature space via convolutional groups, producing features of dimension $\mathbb{R}^{C \times sH \times sW}$. These features are processed through a U-shaped network to capture multi-scale contextual information. In the encoder stage, spatial resolution is progressively reduced while the channel dimension increases, resulting in enriched latent representations. Subsequently, a symmetric decoder restores spatial resolution to reconstruct high-resolution features. Both the encoder and decoder are composed of multiple stacked Mamba Blocks. Each Mamba block consists of an Interactive State Space Model (ISSM), a Feed-forward Network (FFN), and a Cross-Modal Matching Transform (CMT) module. Finally, the decoder outputs are concatenated and refined via convolutional groups to generate a residual map $\boldsymbol{R}$, which is combined with the upsampled depth $\boldsymbol{D}{up}$ through a skip connection~\cite{he2016deep} to produce the final high-resolution depth reconstruction: $\boldsymbol{D}_{hr} = \boldsymbol{D}_{up} + \boldsymbol{R}$.

\subsection{Interactive State Space Model}
The Interactive State Space Model (ISSM) constitutes the core component of our framework. It incorporates a specifically designed Cross-modal Local Scanning (CMLS) mechanism while maintaining linear computational complexity, thereby ensuring high operational efficiency. For the input depth features $\boldsymbol{F}_{depth}$ and RGB features $\boldsymbol{F}_{RGB}$, the process begins with layer normalization, followed by a linear projection and a depth-wise convolution that map the features to the higher-dimensional space. The SiLU activation function is then applied to enhance non-linear representation capability. This procedure can be formulated as:
\begin{align}
\hat{\boldsymbol{F}}_{depth}&=\sigma(\operatorname{DConv}(\operatorname{Linear}(\operatorname{LN}(\boldsymbol{F}_{depth})))), \\
\hat{\boldsymbol{F}}_{RGB}&=\sigma(\operatorname{DConv}(\operatorname{Linear}(\operatorname{LN}(\boldsymbol{F}_{RGB})))),
\end{align}
where $\operatorname{LN}(\cdot)$ denotes the layer normalization, $\operatorname{Linear}(\cdot)$ stands for the liear projection, $\operatorname{DConv}(\cdot)$ is the depth-wise convolution and $\sigma(\cdot)$ represents Silu function. 

Subsequently, the specifically designed CMLS mechanism is applied to $\hat{\boldsymbol{F}}_{depth}$ and $\hat{\boldsymbol{F}}_{RGB}$ to achieve fine-grained semantic interaction and long-range dependency modeling between the depth and RGB modalities. The CMLS mechanism operates through three sequential steps: rearrange, patchify, and S6 scanning. Our observations indicate that after feature extraction, certain feature maps from different modalities exhibit strong semantic correlations. That is, although these features originate from distinct modalities, they attend to highly similar semantic information in specific channels to collectively contribute to high-quality output. To mitigate information decay during long-range modeling, we aim to ensure that these semantically related channel features remain in close proximity. The rearrange step dynamically adjusts the channel order to cluster correlated feature channels, thereby facilitating fine-grained semantic interaction. Furthermore, following previous methods, we introduce a patchify operation to enhance the local inductive bias within the state space model. Finally, these features from depth and RGB are processed by the S6 block to efficiently capture global dependencies with linear complexity. This process is summarized as:
\begin{align}
    \boldsymbol{F}_{g}^{depth},\boldsymbol{F}_{g}^{RGB}=\operatorname{S6}(\operatorname{P}(\operatorname{R}(\hat{\boldsymbol{F}}_{depth},\hat{\boldsymbol{F}}_{RGB}))),
\end{align}
where $\operatorname{R}(\cdot,\cdot)$ is the rearrange operation and $\operatorname{P}(\cdot)$ denotes the patchify operation. $\operatorname{S6}(\cdot)$ block denotes the selective scanning mechanism proposed by Mamba. Finally, the globally enhanced features are processed through layer normalization and a gating mechanism to refine their representational capacity by adaptively emphasizing informative components while suppressing less relevant features. This process can be formulated as:
\begin{align}
    \boldsymbol{F}_{w}^{depth}&=\operatorname{Linear}(\operatorname{LN}(\boldsymbol{F}_{depth})),\\
    \boldsymbol{F}_{w}^{RGB}&=\operatorname{Linear}(\operatorname{LN}(\boldsymbol{F}_{RGB})),\\
    \boldsymbol{F}_{out}^{depth}&=\operatorname{Linear}(\operatorname{LN}(\boldsymbol{F}_{g}^{depth}) \odot \boldsymbol{F}_{w}^{depth}),\\
    \boldsymbol{F}_{out}^{RGB}&=\operatorname{Linear}(\operatorname{LN}(\boldsymbol{F}_{g}^{RGB}) \odot \boldsymbol{F}_{w}^{RGB}),
\end{align}
where $\operatorname{\odot}$ is the element-wise multiplication.

\subsection{Cross-modol Matching Transform}
Following the architectural paradigm established by MetaFormer~\cite{yu2022metaformer}, we incorporate a Feed-Forward Network (FFN) after the ISSM to further enhance non-linear representation capacity and refine local features~\cite{Restormer}. However, this FFN stage remains modality-specific and fails to effectively leverage complementary information across modalities. To address this limitation, we introduce a Cross-Modal Matching Transform (CMMT) module after the FFN. The CMMT module explicitly aligns and fuses semantically similar features across modalities, thereby enhancing inter-modal representation through targeted feature interaction and integration. Given input features $\boldsymbol{F}_{depth} \in \mathbb{R}^{C \times H \times W}$ and $\boldsymbol{F}_{RGB} \in \mathbb{R}^{C \times H \times W}$, we first apply a convolutional layer to enrich their local contextual representations, producing enhanced features $\hat{\boldsymbol{F}}_{depth}$ and $\hat{\boldsymbol{F}}_{RGB}$. Subsequently, a dimensionality transformation is applied by flattening the spatial dimensions, and the similarity matrix $\boldsymbol{M} \in \mathbb{R}^{C \times C}$ between the two features is computed. Based on the $\boldsymbol{M}$, the top-1 most relevant vector is selected and sorted to form a sequence $\boldsymbol{S} \in \mathbb{R}^{C \times 1}$, which indicates the features most correlated with $\boldsymbol{F}_{depth}$. We then select the top-$C/r$ (r $\geq$ 1 means the squeezing factor which controls the squeezing level) features from $\boldsymbol{F}_{RGB}$ to interact with $\boldsymbol{F}_{depth}$, effectively enabling cross-modal fusion and enhancing the representational capacity of the depth features. This process can be formulated as:
\begin{align}
    &\hat{\boldsymbol{F}}_{depth}=\operatorname{Conv}(\boldsymbol{F}_{depth}), \hat{\boldsymbol{F}}_{RGB}=\operatorname{Conv}(\boldsymbol{F}_{RGB}),\\
    &\boldsymbol{M}=\operatorname{Sim}(\hat{\boldsymbol{F}}_{depth}, \hat{\boldsymbol{F}}_{RGB}),\boldsymbol{S}=\operatorname{Sort}(\operatorname{Top_{1}}(\boldsymbol{M})),\\
    &\boldsymbol{F}_{fusion}=[\operatorname{Select}_{C/r}(\hat{\boldsymbol{F}}_{RGB},\boldsymbol{S});\hat{\boldsymbol{F}}_{depth}],\\
    &\boldsymbol{F}_{out}=\operatorname{Conv}(\operatorname{Sig}(\operatorname{Conv}(\boldsymbol{F}_{fusion}))\odot\operatorname{Conv}(\boldsymbol{F}_{fusion})),
\end{align}
where $\operatorname{Sim}(\cdot,\cdot)$ denotes the operation of the computer similarity, which is measured in terms of Euclidean distances. $\operatorname{Select}(\cdot,\cdot)$ and $\operatorname{Sort}$ represent the select features operation and obtain indices value operation. $\operatorname{[\cdot;\cdot]}$ is the concatenation operation and $\operatorname{Sig}$ stands for Sigmoid function. The aforementioned process enhances the depth features and can be formulated as $\operatorname{CMMT}(\boldsymbol{F}_{depth}, \boldsymbol{F}_{RGB})$. Similarly, we apply the same operation to the RGB features to achieve fine-grained refinement, denoted as $\operatorname{CMMT}(\boldsymbol{F}_{RGB}, \boldsymbol{F}_{depth})$. This symmetric processing enables mutual enhancement and optimal alignment between the two modalities.

\begin{figure*}[t!]
    \centering
    \includegraphics[width=.85\linewidth]{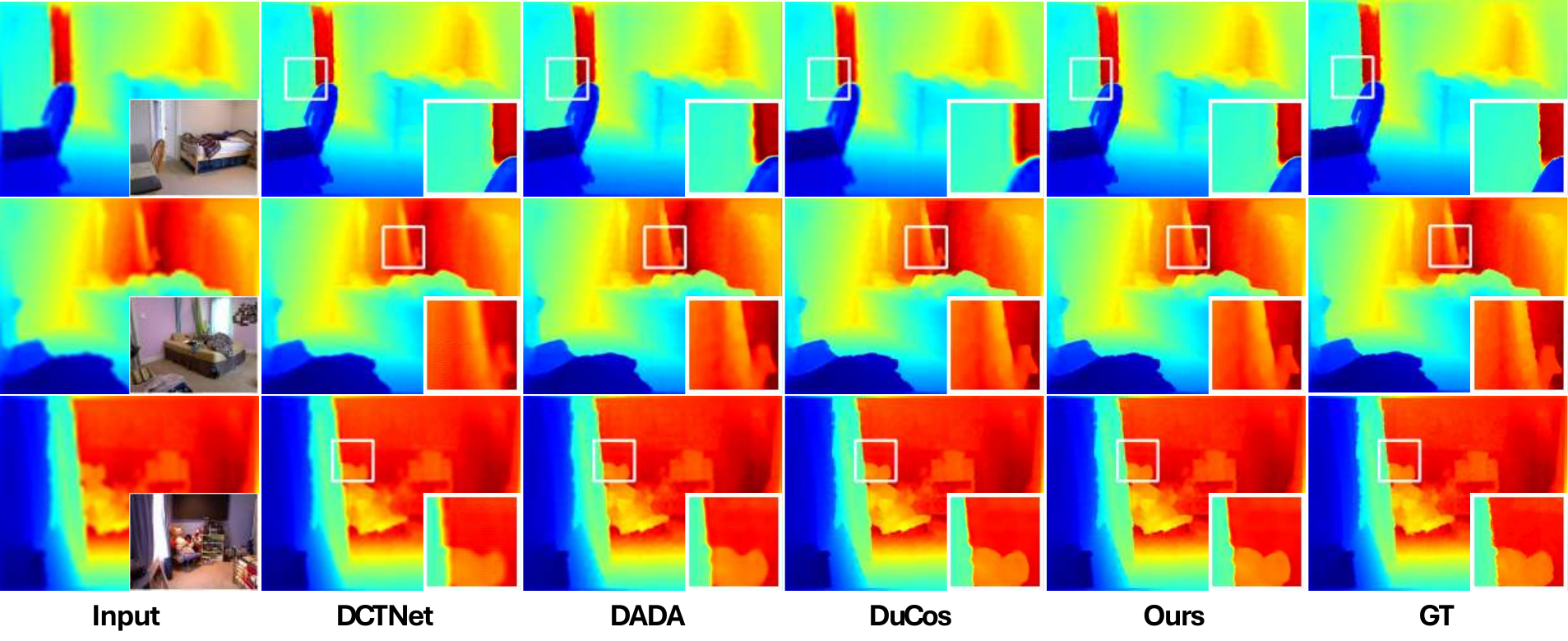}
    \caption{Visual quality comparisons on NYU-v2 dataset. Please zoom in for details.}
    \label{fig:result}
\end{figure*}

\input{tabs/res}

\section{Experiment}
\subsection{Experimental Settings}
\subsubsection{Implementation Details}
We conduct experiments in PyTorch on a NVIDIA GeForce RTX 4090 GPU. During the training, we randomly crop the RGB image and HR depth into $256\times256$. Adam optimizer with an initial learning rate (lr) of $2e^{-4}$ is used to train our network. The channel dimension and the number of training epochs are set to 32 and 200, respectively. The network parameters are optimized using a composite loss function combining L1 loss and Fourier loss, where the weight of the Fourier loss term is empirically set to 0.002 to balance their contributions~\cite{SGNet}. The FFN used in this paper is the Gated-Dconv Feed-Forward Network (GDFN)~\cite{Restormer}.

\subsection{Datasets and Evaluation}
Comprehensive experiments are conducted to evaluate our method across three widely-used benchmarks: NYU-v2~\cite{nyuv2}, Middlebury~\cite{Middlebury1,Middlebury2}, and Lu~\cite{Lu} dataset, under multiple super-resolution scales ($\times4$, $\times8$, $\times16$). All models are trained on the Hypersim~\cite{Hypersim}. Following previous methods~\cite{SGNet,DKN,DCTNet}, we employ the Root Mean Square Error (RMSE), measured in centimeters, as the primary evaluation metric.

\subsection{Comparisons with the State-of-the-art Methods}
We compared our method with several existing approaches, including DJF~\cite{DJF}, DJFR~\cite{DJFR}, CUNet~\cite{CUNet}, FDKN~\cite{DKN}, DKN~\cite{DKN}, FDSR~\cite{FDSR}, DCTNet~\cite{DCTNet}, DADA~\cite{DADA}, and DuCos~\cite{DuCos}. The results are shown in Tab.~\ref{tab:res}. Our method achieved best results across multiple datasets, notably reducing the RMSE by 10.38\% compared to the best method, DuCos, on the NYU-v2 $4\times$ GDSR task. Furthermore, we present the visual comparison results with various methods in Fig.~\ref{fig:result}, where it can be observed that our approach produces results closest to the ground truth.

\input{tabs/abla1}

\subsection{Ablation Studies and Discussions}
To validate the effectiveness of the proposed module, we conducted a series of ablation studies. To ensure fairness, we replaced the proposed module with ResBlock~\cite{he2016deep} instead of directly removing it. It is note worthy that -R/D indicates RGB/Depth as the primary modality.

As summarized in Tab.~\ref{tab:abla1}, removing both ISSM and CMMT leads to the most significant performance degradation, confirming the importance of our designed modules. ISSM aimed to facilitate fine-grained semantic interactions between modalities and global dependency modeling. Its absence led to a significant decline, indicating that semantic-enhanced global modeling is essential, as it provided a substantial performance boost to the model. Similarly, removing the CMMT module also results in performance degradation. We observe that the performance gain with depth as the primary modality was greater than that with RGB, which is quite intuitive, as depth information is our ultimate reconstruction target, while RGB essentially serves as an auxiliary modality. Overall, these components are all essential, and their collective presence yields optimal results.

\section{Conclusion}
In this paper, we propose a novel framework based on the ISSM, which introduces the Mamba architecture with linear computational complexity into the GDSR task for the first time. Through an innovative CMLS mechanism, we consider fine-grained semantic interactions between RGB and depth features when modeling long-range dependencies, effectively alleviating information degradation in global modeling. Additionally, we design the CMMT module to further enhance the representation of each modality from the perspective of feature matching. Extensive experiments confirm the effectiveness of our method.

\bibliographystyle{IEEEtran}
\bibliography{ref}

\end{document}

%% file: tabs/res.tex
\begin{table}[t!]
\centering
\caption{Quantitative comparisons on the several DSR benchmarks. Best and second best values are indicated with \textbf{bold} text and \underline{underlined} text respectively.}
\label{tab:res}
\resizebox{\linewidth}{!}{
\begin{tabular}{c|ccccccccc}
\toprule[0.5mm]
\multirow{2}{*}{Method} & \multicolumn{3}{c}{Middlebury} & \multicolumn{3}{c}{Lu} & \multicolumn{3}{c}{NYU-v2} \\ \cline{2-10} 
                        & ×4       & ×8       & ×16      & ×4     & ×8    & ×16   & ×4      & ×8      & ×16    \\ \hline
DJF~\cite{DJF}                     & 1.93     & 3.09     & 5.50     & 1.93   & 3.58  & 6.53  & 3.60    & 5.56    & 9.82   \\
DJFR~\cite{DJFR}                    & 1.83     & 2.82     & 5.16     & 1.85   & 3.24  & 6.46  & 3.27    & 5.20    & 9.50   \\
CUNet~\cite{CUNet}                   & 1.61     & 2.86     & 4.72     & 1.73   & 2.85  & 5.63  & 3.22    & 5.50    & 8.63   \\
FDKN~\cite{DKN}                    & 1.80     & 2.51     & 4.42     & 2.11   & 2.67  & 5.48  & 3.28    & 4.93    & 7.97   \\
DKN~\cite{DKN}                     & 1.77     & 2.43     & 4.17     & 2.05   & 2.88  & 5.44  & 3.15    & 4.88    & 7.70   \\
FDSR~\cite{FDSR}                    & 1.72     & 2.41     & 3.97     & 1.95   & 2.69  & 5.23  & 2.94    & 4.82    & \textbf{7.29}   \\
DCTNet~\cite{DCTNet}                  & 1.66     & 2.75     & 5.07     & 1.85   & 3.07  & 5.83  & 2.90    & 4.90    & 9.10   \\
DADA~\cite{DADA}                    & 1.82     & 2.77     & 4.11     & 2.12   & 3.76  & 6.19  & 3.08    & 4.83    & 7.99   \\
DuCos~\cite{DuCos}                   &\underline{1.45}  & \underline{2.23}     & \underline{3.96}     & \underline{1.38}   & \textbf{2.67}  & \underline{5.18}  & \underline{2.60}    & \underline{4.61}    & 7.37   \\ \hline
Ours                    & \textbf{1.44}     & \textbf{2.17}     & \textbf{3.61}     & \textbf{1.35}   & 2.69  & \textbf{5.11}  & \textbf{2.33}    & \textbf{4.53}    & \underline{7.31}   \\ 
\bottomrule[0.5mm]
\end{tabular}}
\end{table}

%% file: tabs/abla1.tex
\begin{table}[t!]
\centering
\caption{Ablation studies of proposed modules.}
\label{tab:abla1}
\resizebox{.6\linewidth}{!}{
\begin{tabular}{cccc|c}
\toprule[0.5mm]
ISSM          & CMMT-R     & CMMT-D     & FFN        & RMSE \\ \hline
              &            &            & \checkmark & 3.19 \\
\checkmark    &            &            & \checkmark & 2.55 \\
\checkmark    & \checkmark &            & \checkmark & 2.44 \\
\checkmark    &            & \checkmark & \checkmark & 2.41 \\
              & \checkmark & \checkmark & \checkmark & 2.76 \\
\checkmark    & \checkmark & \checkmark & \checkmark & 2.33 \\ 
\bottomrule[0.5mm]
\end{tabular}}
\end{table}

%% file: ref.bib
@inproceedings{DJF,
  title={Deep joint image filtering},
  author={Li, Yijun and Huang, Jia-Bin and Ahuja, Narendra and Yang, Ming-Hsuan},
  booktitle={European conference on computer vision},
  pages={154--169},
  year={2016},
  organization={Springer}
}

@article{DJFR,
  title={Joint image filtering with deep convolutional networks},
  author={Li, Yijun and Huang, Jia-Bin and Ahuja, Narendra and Yang, Ming-Hsuan},
  journal={IEEE transactions on pattern analysis and machine intelligence},
  volume={41},
  number={8},
  pages={1909--1923},
  year={2019},
  publisher={IEEE}
}

@article{CUNet,
  title={Deep convolutional neural network for multi-modal image restoration and fusion},
  author={Deng, Xin and Dragotti, Pier Luigi},
  journal={IEEE transactions on pattern analysis and machine intelligence},
  volume={43},
  number={10},
  pages={3333--3348},
  year={2020},
  publisher={IEEE}
}

@article{DKN,
  title={Deformable kernel networks for joint image filtering},
  author={Kim, Beomjun and Ponce, Jean and Ham, Bumsub},
  journal={International Journal of Computer Vision},
  volume={129},
  number={2},
  pages={579--600},
  year={2021},
  publisher={Springer}
}

@inproceedings{FDSR,
  title={Towards fast and accurate real-world depth super-resolution: Benchmark dataset and baseline},
  author={He, Lingzhi and Zhu, Hongguang and Li, Feng and Bai, Huihui and Cong, Runmin and Zhang, Chunjie and Lin, Chunyu and Liu, Meiqin and Zhao, Yao},
  booktitle={Proceedings of the ieee/cvf conference on computer vision and pattern recognition},
  pages={9229--9238},
  year={2021}
}

@inproceedings{DCTNet,
  title={Discrete cosine transform network for guided depth map super-resolution},
  author={Zhao, Zixiang and Zhang, Jiangshe and Xu, Shuang and Lin, Zudi and Pfister, Hanspeter},
  booktitle={Proceedings of the IEEE/CVF conference on computer vision and pattern recognition},
  pages={5697--5707},
  year={2022}
}

@inproceedings{DADA,
  title={Guided depth super-resolution by deep anisotropic diffusion},
  author={Metzger, Nando and Daudt, Rodrigo Caye and Schindler, Konrad},
  booktitle={Proceedings of the IEEE/CVF Conference on Computer Vision and Pattern Recognition},
  pages={18237--18246},
  year={2023}
}

@article{DuCos,
  title={DuCos: Duality Constrained Depth Super-Resolution via Foundation Model},
  author={Yan, Zhiqiang and Wang, Zhengxue and Dong, Haoye and Li, Jun and Yang, Jian and Lee, Gim Hee},
  journal={arXiv preprint arXiv:2503.04171},
  year={2025}
}

@inproceedings{DRMPNet,
  title={Delving into Transformer-based Network Architecture for Guided Depth Super-Resolution},
  author={Ye, Xinchen and Zhang, Aokai and Xu, Rui and Li, Haojie},
  booktitle={ICASSP 2025-2025 IEEE International Conference on Acoustics, Speech and Signal Processing (ICASSP)},
  pages={1--5},
  year={2025},
  organization={IEEE}
}

@inproceedings{SUFT,
  title={Symmetric uncertainty-aware feature transmission for depth super-resolution},
  author={Shi, Wuxuan and Ye, Mang and Du, Bo},
  booktitle={Proceedings of the 30th ACM International Conference on Multimedia},
  pages={3867--3876},
  year={2022}
}

@inproceedings{RSAG,
  title={Recurrent structure attention guidance for depth super-resolution},
  author={Yuan, Jiayi and Jiang, Haobo and Li, Xiang and Qian, Jianjun and Li, Jun and Yang, Jian},
  booktitle={Proceedings of the AAAI Conference on Artificial Intelligence},
  volume={37},
  number={3},
  pages={3331--3339},
  year={2023}
}

@inproceedings{SGNet,
  title={Sgnet: Structure guided network via gradient-frequency awareness for depth map super-resolution},
  author={Wang, Zhengxue and Yan, Zhiqiang and Yang, Jian},
  booktitle={Proceedings of the AAAI Conference on Artificial Intelligence},
  volume={38},
  number={6},
  pages={5823--5831},
  year={2024}
}

@inproceedings{Restormer,
  title={Restormer: Efficient transformer for high-resolution image restoration},
  author={Zamir, Syed Waqas and Arora, Aditya and Khan, Salman and Hayat, Munawar and Khan, Fahad Shahbaz and Yang, Ming-Hsuan},
  booktitle={Proceedings of the IEEE/CVF conference on computer vision and pattern recognition},
  pages={5728--5739},
  year={2022}
}

@article{bonetti2018augmented,
  title={Augmented reality and virtual reality in physical and online retailing: A review, synthesis and research agenda},
  author={Bonetti, Francesca and Warnaby, Gary and Quinn, Lee},
  journal={Augmented reality and virtual reality},
  pages={119--132},
  year={2018},
  publisher={Springer}
}

@inproceedings{wang2019pseudo,
  title={Pseudo-lidar from visual depth estimation: Bridging the gap in 3d object detection for autonomous driving},
  author={Wang, Yan and Chao, Wei-Lun and Garg, Divyansh and Hariharan, Bharath and Campbell, Mark and Weinberger, Kilian Q},
  booktitle={Proceedings of the IEEE/CVF conference on computer vision and pattern recognition},
  pages={8445--8453},
  year={2019}
}

@article{chowdhary20193d,
  title={3D object recognition system based on local shape descriptors and depth data analysis},
  author={Chowdhary, Chiranji L},
  journal={Recent Patents on Computer Science},
  volume={12},
  number={1},
  pages={18--24},
  year={2019},
  publisher={Bentham Science Publishers}
}

@inproceedings{Mamba,
  title={Mamba: Linear-time sequence modeling with selective state spaces},
  author={Gu, Albert and Dao, Tri},
  booktitle={First Conference on Language Modeling},
  year={2024}
}

@inproceedings{he2016deep,
  title={Deep residual learning for image recognition},
  author={He, Kaiming and Zhang, Xiangyu and Ren, Shaoqing and Sun, Jian},
  booktitle={Proceedings of the IEEE conference on computer vision and pattern recognition},
  pages={770--778},
  year={2016}
}

@inproceedings{yu2022metaformer,
  title={Metaformer is actually what you need for vision},
  author={Yu, Weihao and Luo, Mi and Zhou, Pan and Si, Chenyang and Zhou, Yichen and Wang, Xinchao and Feng, Jiashi and Yan, Shuicheng},
  booktitle={Proceedings of the IEEE/CVF conference on computer vision and pattern recognition},
  pages={10819--10829},
  year={2022}
}

@inproceedings{nyuv2,
  title={Indoor segmentation and support inference from rgbd images},
  author={Silberman, Nathan and Hoiem, Derek and Kohli, Pushmeet and Fergus, Rob},
  booktitle={European conference on computer vision},
  pages={746--760},
  year={2012},
  organization={Springer}
}

@inproceedings{Middlebury1,
  title={Evaluation of cost functions for stereo matching},
  author={Hirschmuller, Heiko and Scharstein, Daniel},
  booktitle={2007 IEEE conference on computer vision and pattern recognition},
  pages={1--8},
  year={2007},
  organization={IEEE}
}

@inproceedings{Middlebury2,
  title={Learning conditional random fields for stereo},
  author={Scharstein, Daniel and Pal, Chris},
  booktitle={2007 IEEE conference on computer vision and pattern recognition},
  pages={1--8},
  year={2007},
  organization={IEEE}
}

@inproceedings{Lu,
  title={Depth enhancement via low-rank matrix completion},
  author={Lu, Si and Ren, Xiaofeng and Liu, Feng},
  booktitle={Proceedings of the IEEE conference on computer vision and pattern recognition},
  pages={3390--3397},
  year={2014}
}

@inproceedings{Hypersim,
  title={Hypersim: A photorealistic synthetic dataset for holistic indoor scene understanding},
  author={Roberts, Mike and Ramapuram, Jason and Ranjan, Anurag and Kumar, Atulit and Bautista, Miguel Angel and Paczan, Nathan and Webb, Russ and Susskind, Joshua M},
  booktitle={Proceedings of the IEEE/CVF international conference on computer vision},
  pages={10912--10922},
  year={2021}
}

@article{peng2025directing,
  title={Directing mamba to complex textures: An efficient texture-aware state space model for image restoration},
  author={Peng, Long and Di, Xin and Feng, Zhanfeng and Li, Wenbo and Pei, Renjing and Wang, Yang and Fu, Xueyang and Cao, Yang and Zha, Zheng-Jun},
  journal={arXiv preprint arXiv:2501.16583},
  year={2025}
}

@article{he2024multi,
  title={Multi-scale representation learning for image restoration with state-space model},
  author={He, Yuhong and Peng, Long and Yi, Qiaosi and Wu, Chen and Wang, Lu},
  journal={arXiv preprint arXiv:2408.10145},
  year={2024}
}

@article{xia2024s3mamba,
  title={S3mamba: Arbitrary-scale super-resolution via scaleable state space model},
  author={Xia, Peizhe and Peng, Long and Di, Xin and Pei, Renjing and Wang, Yang and Cao, Yang and Zha, Zheng-Jun},
  journal={arXiv preprint arXiv:2411.11906},
  volume={6},
  year={2024}
}

@inproceedings{huang2024localmamba,
  title={Localmamba: Visual state space model with windowed selective scan},
  author={Huang, Tao and Pei, Xiaohuan and You, Shan and Wang, Fei and Qian, Chen and Xu, Chang},
  booktitle={European Conference on Computer Vision},
  pages={12--22},
  year={2024},
  organization={Springer}
}

@article{yurtsever2020survey,
  title={A survey of autonomous driving: Common practices and emerging technologies},
  author={Yurtsever, Ekim and Lambert, Jacob and Carballo, Alexander and Takeda, Kazuya},
  journal={IEEE access},
  volume={8},
  pages={58443--58469},
  year={2020},
  publisher={IEEE}
}

@book{burdea2003virtual,
  title={Virtual reality technology},
  author={Burdea, Grigore C and Coiffet, Philippe},
  year={2003},
  publisher={John Wiley \& Sons}
}

@inproceedings{riegler2016atgv,
  title={Atgv-net: Accurate depth super-resolution},
  author={Riegler, Gernot and R{\"u}ther, Matthias and Bischof, Horst},
  booktitle={European conference on computer vision},
  pages={268--284},
  year={2016},
  organization={Springer}
}

@inproceedings{song2016deep,
  title={Deep depth super-resolution: Learning depth super-resolution using deep convolutional neural network},
  author={Song, Xibin and Dai, Yuchao and Qin, Xueying},
  booktitle={Asian conference on computer vision},
  pages={360--376},
  year={2016},
  organization={Springer}
}

@inproceedings{voynov2019perceptual,
  title={Perceptual deep depth super-resolution},
  author={Voynov, Oleg and Artemov, Alexey and Egiazarian, Vage and Notchenko, Alexander and Bobrovskikh, Gleb and Burnaev, Evgeny and Zorin, Denis},
  booktitle={Proceedings of the ieee/cvf international conference on computer vision},
  pages={5653--5663},
  year={2019}
}

@article{xu2026fusion,
  title={Fusion Requires Interaction: A Hybrid Mamba-Transformer Architecture for Deep Interactive Fusion of Multi-modal Images},
  author={Xu, Wenxiao and Wu, Chen and Yin, Qiyuan and Wang, Ling and Zheng, Zhuoran and Huang, Daqing},
  journal={Expert Systems with Applications},
  pages={131309},
  year={2026},
  publisher={Elsevier}
}

@article{wu2025ultra,
  title={Ultra-High-Definition Image Restoration via High-Frequency Enhanced Transformer},
  author={Wu, Chen and Wang, Ling and Zheng, Zhuoran and Jiang, Weidong and Cui, Yuning and Xia, Jingyuan},
  journal={IEEE Transactions on Circuits and Systems for Video Technology},
  year={2025},
  publisher={IEEE}
}

@inproceedings{wang2025dap,
  title={Dap-led: Learning degradation-aware priors with clip for joint low-light enhancement and deblurring},
  author={Wang, Ling and Wu, Chen and Wang, Lin},
  booktitle={2025 IEEE International Conference on Robotics and Automation (ICRA)},
  pages={15791--15797},
  year={2025},
  organization={IEEE}
}

@article{wu2025adaptive,
  title={Adaptive feature selection modulation network for efficient image super-resolution},
  author={Wu, Chen and Wang, Ling and Su, Xin and Zheng, Zhuoran},
  journal={IEEE Signal Processing Letters},
  year={2025},
  publisher={IEEE}
}

@article{chen2025mixnet,
  title={MixNet: Efficient global modeling for ultra-high-definition image restoration},
  author={Chen, Wu and Sun, Shuning and Zhang, Yu and Zheng, Zhuoran},
  journal={Neurocomputing},
  pages={131130},
  year={2025},
  publisher={Elsevier}
}
